\documentclass[10pt,journal,cspaper,compsoc]{IEEEtran}

\usepackage{amsmath,epsfig}    % need for subequations
\usepackage{graphicx}   % need for figures
\usepackage{subfig}
\usepackage{amssymb}

\usepackage{fmtcount}

\providecommand{\bw}   {\textbf{w}} 
\providecommand{\bW}   {\textbf{W}}

\providecommand{\bV}   {\textbf{V}}
\providecommand{\bI}   {\textbf{I}} 
\providecommand{\bu}   {\textbf{u}} 
\providecommand{\bv}   {\textbf{v}} 
\providecommand{\bb}   {\textbf{b}} 
\providecommand{\bff}   {\textbf{f}} 
\DeclareMathSymbol{\R}{\mathalpha}{AMSb}{"52}

\begin{document}

%\begin{frontmatter}

\title{Max-Margin Stacking and \\Sparse Regularization for \\Linear Classifier Combination and Selection}

\author{Mehmet~Umut~Sen,~\IEEEmembership{Member,~IEEE,}
        Hakan~Erdogan,~\IEEEmembership{Member,~IEEE} % <-this % stops a space
\thanks{The authors are with Sabanci University, Istanbul Turkey. Email addresses: umutsen@sabanciuniv.edu, haerdogan@sabanciuniv.edu.}
}

%\author{Mehmet Umut Sen}
%\ead{umutsen@sabanciuniv.edu}

%\author{Hakan Erdogan}
%\ead{haerdogan@sabanciuniv.edu}
%\address{Sabanci University, 34956, Istanbul, Turkey}
%\IEEEcompsoctitleabstractindextext{%

\maketitle

\begin{abstract}
The main principle of stacked generalization (or Stacking) is using a second-level generalizer to combine the outputs of base classifiers in an ensemble. In this paper, we investigate different combination types under the stacking framework; namely weighted sum (WS), class-dependent weighted sum (CWS) and linear stacked generalization (LSG). For learning the weights, we propose using regularized empirical risk minimization with the hinge loss. In addition, we propose using group sparsity for regularization to facilitate classifier selection. We performed experiments using two different ensemble setups with differing diversities on 8 real-world datasets. Results show the power of regularized learning with the hinge loss function. Using sparse regularization, we are able to reduce the number of selected classifiers of the diverse ensemble without sacrificing accuracy. With the non-diverse ensembles, we even gain accuracy on average by using sparse regularization.
\end{abstract}

\begin{keywords}
classifier combination, classifier selection, regularized empirical risk minimization, hinge loss, group sparsity
\end{keywords}
%}

%\end{frontmatter}

\section{Introduction}
Classifier ensembles aim to increase efficiency of classifier systems in terms of accuracy at the expense of increased complexity and they are shown to obtain greater performance than single-expert systems for a broad range of applications. Among all theoretical and practical reasons to prefer using ensembles, which are categorized as \textit{statistical}, \textit{computational} and \textit{representational} in \cite{Dietterich00ensemblemethods}, the most important ones are the statistical reasons. Since we are looking for the generalization performance (error in the test data) in pattern recognition problems, it is often very difficult to find the ``perfect classifier'', but by combining multiple classifiers probability of getting closer to the perfect classifier is increased. An ensemble may not always beat the performance of the best single classifier obtained, but it will surely decrease the variance of the classification error. Some other reasons besides statistical reasons can be found in \cite{Dietterich00ensemblemethods,Polikar_2006}.

The straightforward method to obtain an ensemble is using different classifier types or different parameters. Also training base classifiers with different subsets or samplings of data or features is used to obtain ensembles which will result in more diverse ensembles. There are different measures of diversity of an ensemble, but diversity simply means that base classifiers make errors on different examples. Diverse ensembles result in  better performance with a reasonable combiner. In this work, we are not interested in the methods of obtaining the ensemble, but we investigate various linear combination types for a given set of base classifiers.

Base classifiers produce either label outputs or continuous valued outputs. For the former, combiners like majority voting or weighted majority voting are used. In the latter case, base classifiers produce continuous scores for each class that represent the degree of support for each class. They can be interpreted as confidences in the suggested labels or estimates of the posterior probabilities for the classes \cite{kuncheva04cpc}. Former thinking is more reasonable since for most of the classifier types, support values may not be very close to the actual posterior probabilities even if the data is dense, because classifiers generally do not try to estimate the posterior probabilities, but try to classify the data instances correctly so they usually only try to force the true class' score to be the maximum. In this paper, we deal with the combination of continuous valued outputs. 

Combination rules can be grouped into trainable vs. non-trainable (or supervised vs. unsupervised). Simple average (mean), product, trimmed mean, minimum, maximum and median rules are some examples to non-trainable combiners. Learning the combiner from training data is shown to give better accuracy than non-trainable combiners. Among trainable combiners, such as stacked generalization (Stacking) \cite{wolpert92sg}, Decision Templates \cite{kuncheva04cpc} and Dempster-Shafer Combination \cite{Rogova94}; stacked generalization is deeply investigated and analyzed in the literature \cite{wolpert92sg,ting99isg,LeBlanc93,Ting97stackingbagged,Seewald2002,SCANN99,Dzeroski02,sill09,Reid_2009,Todorovski00,Ledezma10}. The idea of Stacking is to use the confidence scores that are obtained from base classifiers as attributes in a new training set with the original class labels and training a meta-classifier (This classifier is called level-1 generalizer in \cite{wolpert92sg}) with this new dataset. Considering the speed and complexity advantage of linear meta-classifiers over non-linear ones, they are usually preferred in the literature. When initially introduced, stacking is used to combine the class predictions of the base classifiers \cite{wolpert92sg}. Ting \& Witten used confidence scores of base classifiers as input features and improved stacking's performance \cite{ting99isg,Ting97stackingbagged}. Merz used stacking and correspondence analysis to model the relationship between the learning examples and their classification by a collection of learned models and used nearest neighbor classifier as meta learner. Dzeroski \& Zenko used multi-response model trees as the meta-learner \cite{Dzeroski02}. Seewald introduced StackingC, which improves Stacking's performance further and reduces the computational cost by introducing class-conscious combination. \cite{Seewald2002}. Sill, incorporated meta-features with the posterior scores of base classifiers to improve accuracy \cite{sill09}. Ledezma, used genetic algorithms to search for good Stacking configurations \cite{Ledezma10}. Tang, re-ranked all possible class labels according to the scores and obtained a learner which outperforms all base classifiers \cite{Tang10}.

Since training the base classifiers and the combiner with the same data samples will result in overfitting, a sophisticated cross-validation is applied to obtain the training data of the combiner (level-1 data). This procedure, called internal cross-validation, is described in section \ref{sec:stacking}. After obtaining level-1 data, there are two main problems remaining for a linear combination: (1.) Which type of combination method should be used? (2.) Given a combination type, how should we learn the parameters of the combiner? For the former problem, Ueda \cite{ueda00olc} defined three linear combination types namely type-1, type-2 and type-3. In this paper, we use the descriptive names weighted sum (WS), class-dependent weighted sum (CWS) and linear stacked generalization (LSG) for these types of combinations respectively and investigate all of them. In \cite{LeBlanc93,Ting97stackingbagged}, LSG is used and CWS combination is proposed in \cite{ting99isg}. For the latter problem, Ting \& Witten proposed a multi-response linear regression algorithm for learning the weights \cite{ting99isg}. Ueda in \cite{ueda00olc} proposed using minimum classification error (MCE) criterion for estimating optimal weights, which increased the accuracies. MCE criterion is an approximation to the zero-one loss function which is not convex, so finding a global optimizer is not always possible. Ueda derived algorithms for different types of combinations with MCE loss using stochastic gradient methods. Both of these studies ignored ``regularization'' which has a huge effect on the performance, especially if the number of base classifiers is large. Reid \& Grudic in \cite{Reid_2009} regularized the standard linear least squares estimation of the weights with CWS and improved the performance of stacking. They applied $l_2$ norm penalty, $l_1$ norm penalty and combination of the two (elastic net regression). In this work, we propose maximum margin algorithms for learning the optimal weights. We work with the regularized empirical risk minimization framework \cite{Lecun06atutorial} and use the hinge loss function with $l_2$ regularization, which corresponds to the support vector machines (SVM). We do not derive algorithms for the solutions of the minimization problems, but state-of-the-art solutions of SVM in the literature can be modified for our problem. 

Another issue, recently addressed in \cite{Zhang201197}, is combination with a sparse weight vector so that we do not use all of the ensemble. Since we do not have to use classifiers which have zero weight on the test phase, overall test time will be much less. Zhang formulated this problem as a linear programming problem for only the WS combination type \cite{Zhang201197}. Reid used $l_1$ norm regularization for CWS combination \cite{Reid_2009}. In this paper, we investigate sparsity issues for all three combination types: WS, CWS and LSG. We use both $l_1$ norm and $l_1-l_2$ norm for regularization in the objective function for CWS and LSG. Latter regularization results in group sparsity, which is deeply investigated and successfully applied to various problems recently.

Throughout the paper, we used $m$ for the classifier subscript, $n$ for the class subscript, $i$ for the data instance subscript, $M$, $N$ and $I$ for the number of classifiers, classes and data instances respectively. Datapoint subscript $i$ is sometimes dropped for simplicity. In section \ref{sec:stacking} we explain the cross-validation technique used in stacked generalization. In section \ref{sec:combinations}, we define the classifier combination problem formally and define three different combination types used in the literature, namely WS, CWS and LSG. In section \ref{sec:learning}, we explain how the weights are learned using regularized empirical risk minimization framework with hinge loss and a regularization function. In section \ref{sec:sparse}, we define sparse regularization functions to enforce classifier selection. In section \ref{sec:exp}, we describe the experiment setups we build. In section \ref{sec:res}, we show the results of our experiments and discuss them.

\section{Stacked Generalization}
\label{sec:stacking}
A novel approach has been introduced in 1992 known as stacked generalization or stacking \cite{wolpert92sg}. The basic idea is applying a meta-level (or level-1) generalizer to the outputs of base classifiers (or level-0 classifiers). For training the level-1 classifier, we need the confidence scores (Level-1 Data) of the training data, but training the combiner with the same data instances which are used for training the base classifiers will lead to overfitting the database and eventually result in poor generalization performance. Stacking deals with this problem by a sophisticated cross-validation method (internal CV), in which training data is divided into $k$ parts and each part of the data is tested with the base classifiers that are trained with the other $k-1$ parts of data. So at the end, each training instance's score is obtained from the base classifiers whose training data does not contain that particular instance. This procedure is repeated for each base classifier in the ensemble. We apply this procedure for the three different linear combination types. Wolpert combined only the class predictions with this framework, Ting \& Witten improved the performance of stacking by combining continuous valued predictions \cite{ting99isg}.

\section{Combination Types}
\label{sec:combinations}

\subsection{Problem Formulation}
In the classifier combination problem, input to the combiner are the posterior scores belonging to different classes obtained from the base classifiers. Let $p_m^n$ be the posterior score of class $n$ obtained from classifier $m$ for any data instance. Let $\textbf{p}_m = [p_m^1,p_m^2,\ldots,p_m^N]^T$, then the input to the combiner is $\textbf{f}=[\textbf{p}_1^T,\textbf{p}_2^T,\ldots,\textbf{p}_M^T]^T$, where $N$ is the number of classes and $M$ is the number of classifiers. Outputs of the combiner are $N$ different scores representing the degree of support for each class. Let $r^n$ be the combined score of class $n$  and let $\textbf{r} = [r^1, \ldots, r^N]^T$; then in general the combiner is defined as a function $g: \R^{MN} \rightarrow \R^N$ such that $\textbf{r} = g(\textbf{f})$. Let $I$ be the number of training data instances, $\textbf{f}_i$ contain the scores for training data point $i$ obtained from base classifiers with stacking and $y_i$ be the corresponding class label; then our aim is to learn the $g$ function using data $\{(\textbf{f}_i,y_i)\}_{i=1}^I$. On the test phase, label of a data instance is assigned as follows:
\begin{equation}
\hat{y} = \operatorname*{arg\,max}_{n \in [N]} r^n,
\end{equation}
where $[N] = \{1,\ldots,N\}$. Among combination types, linear ones are shown to be powerful for the classifier combination problem. For linear combiners, $g$ function has the following form:
\begin{equation}
\label{eq:linComb}
g(\textbf{f}) = \bW\textbf{f} + \textbf{b}.
\end{equation}
In this case, we aim to learn the elements of $\bW\in\R^{N\times MN}$ and $\textbf{b}\in\R^N$. So, the number of parameters to be learned is $MN^2+N$. This type of combination is the most general form of linear combiners and called type-3 combination in \cite{ueda00olc}. In the framework of stacking, we call it linear stacked generalization (LSG) combination. One disadvantage of this type of combination is that, since the number of parameters is high, learning the combiner takes a lot of time and may require a large amount of training data. To overcome this disadvantage, simpler but still strong combiner types are introduced with the help of the knowledge that $p_m^n$ is the posterior score of  class $n$. We call these methods weighted sum (WS) rule and class-dependent weighted sum (CWS) rule. These types are categorized as class-conscious combinations in \cite{kuncheva04cpc}.

\subsection{Linear Combination Types} In this section, we describe and analyze three combination types, namely \textit{weighted sum} rule (WS), \textit{class-dependent weighted sum} rule (CWS) and \textit{linear stacked generalization} (LSG) where LSG is already defined in (\ref{eq:linComb}).

\subsubsection{Weighted Sum Rule}
In this type of combination, each classifier is given a weight, so there are totally $M$ different weights. Let $u_m$ be the weight of classifier $m$, then the final score of class $n$ is estimated as follows:
\begin{equation}
r^n = \sum_{m=1}^M{u_m p_m^n} = \bu^T\bff^n  \makebox[25pt]{,  }  n=1,\ldots,N,
\end{equation}
where  $\bff^n$ contains the scores of class $n$: $\bff^n = [p_1^n,\ldots,p_M^n]^T$ and $\bu = [u_1,\ldots,u_M]^T$. For the framework given in (\ref{eq:linComb}), WS combination can be obtained by letting $\textbf{b} = 0$ and $\bW$ to be the concatenation of constant diagonal matrices:
\begin{equation}
\label{eq:type1_W}
\bW = [u_1 \bI_N | \ldots |u_M \bI_N],
\end{equation}
where $\bI_N$ is the $N\times N$ identity matrix. We expect to obtain higher weights for stronger base classifiers after learning the weights from the database. 

\subsubsection{Class-Dependent Weighted Sum Rule}
The performances of base classifiers may differ for different classes and it may be better to use a different weight distribution for each class. We call this type of combination CWS rule. Let $v_m^n$ be the weight of classifier $m$ for class $n$, then the final score of class $n$ is estimated as follows:
\begin{equation}
r^n = \sum_{m=1}^M{v_m^n p_m^n} = \bv_n^T\bff^n  \makebox[25pt]{,  } n=1,\ldots,N,
\end{equation}
where $\bv_n = [v_1^n,\ldots,v_M^n]^T$. There are $MN$ parameters in a CWS combiner. For the framework given in (\ref{eq:linComb}), CWS combination can be obtained by letting $\textbf{b} = 0$ and $\bW$ to be the concatenation of diagonal matrices; but unlike in WS, diagonals are not constant:
\begin{equation}
\bW = [\bW_1 | \bW_2| \ldots |\bW_M],
\end{equation}
where $\bW_m \in \R^{N\times N}$ are diagonal for $m=1,\ldots ,M$.

\subsubsection{Linear Stacked Generalization}
This type of combination is the most general form of supervised linear combinations and is already defined in (\ref{eq:linComb}). With LSG, score of class $n$ is estimated as follows:
\begin{equation}
r^n = \bw_n^T\bff + b_n \makebox[25pt]{,  } n=1,\ldots,N,
\end{equation}
where $\bw_n \in \R^{MN}$ is the $n^{th}$ row of $\bW$ and $b_n$ is the $n^{th}$ element of $\textbf{b}$. LSG can be interpreted as feeding the base classifiers' outputs to a linear multi-class classifier as a new set of features. This type of combination may result in overfitting to the database and may give lower accuracy then WS and CWS combination when there is not enough data. From this point of view, WS and CWS combination can be treated as regularized versions of LSG. A crucial disadvantage of LSG is that the number of parameters to be learned is $MN^2 + N$ which will result in a long training period. 

There is not a single superior one among these three combination types since results are shown to be data dependent \cite{erdogan10icpr}. A convenient way of choosing the combination type is selecting the one that gives the best performance in cross-validation. 

\section{Learning the Combiner}
\label{sec:learning}
We use the regularized empirical risk minimization (RERM) framework \cite{Lecun06atutorial} for learning the weights. In this framework, learning is formulated as an unconstrained minimization problem and the objective function consists of a summation of empirical risk function over data instances and a regularization function. Empirical risk is obtained as a sum of ``loss'' values obtained from each sample. Different choices of loss functions and regularization functions correspond to different classifiers. Using hinge loss function with $l_2$ norm regularization is equivalent to support vector machines (SVM). It has been shown in studies that the hinge loss function yields much better classification performance as compared to the least-squares(LS) loss function in general. Earlier classifier combination literature uses LS loss function \cite{ting99isg,Ting97stackingbagged,Reid_2009}, which is suboptimal as compared to the hinge loss that we promote and use in this paper. Using least-squares with $l_2$ regularization is equivalent to applying least-squares support vector machine (LS-SVM) \cite{suykens99}. We use an adaptation of SVM in multiclass problems defined in \cite{Crammer01}. With this adaptation, we find the linear separating hyper-plane that maximizes the margin between true class and the most offending wrong class. For LSG, we have the following objective function:
\begin{equation}
\label{eq:rerm}
\phi_{LSG}(\bW,\bb) = \frac{1}{I}\sum_{i=1}^I{(1- r_i^{y_i}  + \max_{n\neq y_i}{r_i^n)_+}} + \lambda  R_{LSG}(\bW),
\end{equation}
where $R_{LSG}(\bW)$ is the regularization function, $(x)_+ = \max(0,x)$ and the posterior score of data instance $i$ for class $n$, $r^n_i$, is given as follows:
\begin{equation}
r_i^n = \bw_n^T\bff_i + b_n.
\end{equation}

$\lambda \in \R$ in (\ref{eq:rerm}) is the regularization parameter which is usually learned by cross validation.
Objective function given in (\ref{eq:rerm}) encourages the distance between the true class' score and the most offending wrong class' score to be larger than one. A conventional regularization function is the Frobenius norm of $\bW$: 
\begin{equation}
R_{LSG}(\bW) = ||\bW||_F = \sum_{n=1}^{N}{||\bw_n||_2^2},
\end{equation}
Equation (\ref{eq:rerm}) is given for LSG but it can be modified for other types of combinations using the unifying framework described in \cite{erdogan10icpr}. But we also give objective functions for WS and CWS explicitly. Objective function of WS is as follows:
\begin{equation}
\label{eq:rermWS}
\phi_{WS}(\bu) = \frac{1}{I}\sum_{i=1}^I{(1-\bu^T\bff_i^{y_i} + \max_{n\neq y_i}{(\bu^T\bff_i^n)})_+} + \lambda R_{WS}(\bu).
\end{equation}
For regularization, we use $l_2$ norm of $\bu$: $R_{WS} = ||\bu||_2$. For CWS, we have the following objective function:
\begin{equation}
\label{eq:rermCWS}
\phi_{CWS}(\bV) = \frac{1}{I}\sum_{i=1}^I{(1-\bv_{y_i}^T\bff_i^{y_i}+\max_{n\neq y_i} {(\bv_n^T\bff_i^n)})_+} + \lambda R_{CWS}(\bV),
\end{equation}
where $\bV \in \R^{M\times N}$ contains the weights for different classes: $\bV=[\bv_1,\ldots,\bv_N]$. As for LSG, conventional regularization function for CWS is the Frobenious norm of $\bV$: $R_{CWS}(\bV) = ||\bV||_F$.

\section{Sparse Regularization}
\label{sec:sparse}
In this section, we define a set of regularization functions for enforcing sparsity on the weights so that the resulting combiner will not use all the base classifiers leading to a shorter test time. This method can be seen as a classifier selection algorithm, but here classifiers are selected automatically and we cannot determine the number of selected classifiers beforehand. But we can lower this number by increasing the weight of the regularization function ($\lambda$), and vice versa. With sparse regularization, $\lambda$ has two main effects on the resulting combiner. First, it will determine how much the combiner should fit the data. Decreasing $\lambda$ results in more fitting the training data and decreasing it too much results in overfitting, on the other hand, increasing it too much prevents the combiner to learn from the data and the accuracy drops dramatically. Second, as mentioned before, it will determine the number of selected classifiers. As $\lambda$ increases, the number of selected classifiers decreases.

\subsection{Regularization with the $l_1$ Norm}
The most successful approach for inducing sparsity is using the $l_1$ norm of the weight vector for WS. For CWS and LSG, in which the combiner consists of matrices, we can concatenate the weights in a vector and take the $l_1$ norm or equivalently we can take the $l_1-l_1$ norm of the weight matrices. We have the following sparse regularization functions for WS, CWS and LSG respectively:
\begin{equation}
R_{WS}(\bu) =  ||\bu||_1,
\end{equation}
\begin{equation}
R_{CWS}(\bV) =  ||\bV||_{1,1} = \sum_{n=1}^N{||\bv_n||_1},
\end{equation}
\begin{equation}
R_{LSG}(\bW) =  ||\bW||_{1,1} = \sum_{n=1}^N{||\bw_n||_1}.
\end{equation}
If all weights of a classifier are zero, that classifier will be eliminated and we do not have to use that base classifier for a test instance, so that testing will be faster. But the problem with $l_1$-norm regularizations for CWS and LSG is that we are not able to use all the information from a selected base classifier, because a classifier may receive both zero and non-zero weights. To overcome this problem, we propose to use group sparsity, as explained in the next section.

\subsection{Regularization with Group Sparsity }
\label{sec:group}
We define another set of regularization functions which are embedded by group sparsity for LSG and CWS to enforce classifier selection. 
%Even though classifier selection may result in better accuracy, the motivation for embeding group lasso is not increasing accuracy but decreasing test time since the number of classifiers which will be used on the test phase will be much less. 
The main principle of the group sparsity is enforcing all elements that belong to a group to be zero altogether. Grouping of the elements are done before learning. In classifier combination, posterior scores obtained from each base classifier form a group. The following regularization function yields group sparsity for LSG:
\begin{equation}
\label{eq:3g}
R_{LSG}(\bW) = \sum_{m=1}^M{||\bW_m||_F}.
\end{equation}
 For CWS, we use the following regularization:
\begin{equation}
R_{CWS}(\bV) = ||\bV||_{1,2} = \sum_{m=1}^M{||\bv^m||_2},
\end{equation}
where $\bv^m$ is the $m^{th}$ row of $\bV$, so it contains the weights of the classifier $m$. After the learning process, the elements of $\bv^m$ for any $m$ are either all zero or all non-zero. This leads to better performance than $l_1$ regularization for automatic classifier selection, as we show in section \ref{sec:res}. In the next section, we describe the setup of the experiments.

\section{Experimental Setups}
\label{sec:exp}
We have performed extensive experiments in eight real-world datasets from the UCI repository \cite{uci}. For a summary of the characteristics of the datasets, see Table \ref{table:datasets}. In order to obtain statistically significant results, we applied 5x2 cross-validation \cite{Dietterich98} which is based on 5 iterations of 2-fold cross-validation (CV). In this method, for each CV, data is randomly split into two stacks as training and testing resulting in overall 10 stacks for each database. 

We constructed two ensembles which differ in their diversity. In the first ensemble, we construct 10 different subsets randomly which contains 80\% of the original data. Then, 13 different classifiers are trained with each subset resulting in a total of 130 base classifiers. We used PR-Tools \cite{prtools} and Libsvm toolbox \cite{CC01a} for obtaining the base classifiers. These 13 different classifiers are: normal densities based linear classifier, normal densities based quadratic classifier, nearest mean classifier, k-nearest neighbor classifier, polynomial classifier, general kernel/dissimilarity based classification, normal densities based classifier with independent features, parzen classifier, binary decision tree classifier, linear perceptron, SVM with linear kernel, polynomial kernel and radial basis function (RBF) kernel. We used default parameters of the toolboxes. In the second ensemble setup, we trained a total of 154 SVM's with different kernel functions and parameters. Latter method produces less diverse base classifiers with respect to the former one. Training data of the combiner is obtained by 4-fold stacked generalization. For each stack in $5 \times 2$ CV, 2-fold CV is used to obtain the optimal $\lambda$ in the regularization function, i.e. $\lambda$ which gives the best average accuracy in CV \footnote{We searched for $\lambda$ in $\{ 10^{-11}, 10^{-9}, 10^{-7}, 10^{-5}, 10^{-3}, 0.005,\\ 0.01, 0.05, 0.1, 0.5, 1, 10 \}$}. For the minimization of the objective functions, we used the CVX-toolbox \cite{cvx}. We use the Wilcoxon signed-rank test for identifying statistical significance of the results with  one-tailed significant level $\alpha = 0.05$ \cite{Demsar06}.

\addtocounter{footnote}{1}

\begin{table}[htb]
\caption{Properties of the data sets used in the experiments  }
\centering
\begin{tabular}{|c|c|c|c|}
\hline 
DB & \# of Instances & \# of classes & \# of features\\
\hline \hline
Segment $^{\decimal{footnote}}$ \addtocounter{footnote}{1} & 2310 & 7 & 19 \\
\hline
Waveform $^{\decimal{footnote}}$ \addtocounter{footnote}{1} & 5000 & 3 & 21\\
\hline
Robot $^{\decimal{footnote}}$ \addtocounter{footnote}{1} & 5456&4 &24\\
\hline
Statlog $^{\decimal{footnote}}$ \addtocounter{footnote}{1} & 846 & 4& 18 \\
\hline
Vowel $^{\decimal{footnote}}$ \addtocounter{footnote}{1} &990 &11 &10\\
\hline
Wine  & 178 & 3 & 13\\
\hline
Yeast & 1484 & 9 & 8 \\
\hline
Steel $^{\decimal{footnote}}$ & 1941 & 7 & 27 \\
\hline
\end{tabular}
 %\footnotetext[2]{The full name of \textit{Robot} dataset is Wall-Following Robot Navigation Data}
\label{table:datasets}
\end{table}

\addtocounter{footnote}{-5}
\footnotetext[\value{footnote}]{The full name of \textit{Segment} dataset is \textit{``Image Segmentation''}}
\addtocounter{footnote}{1}
\footnotetext[\value{footnote}]{The full name of \textit{Waveform} dataset is \textit{``Waveform Database Generator (Version 1)''}}
\addtocounter{footnote}{1}
\footnotetext[\value{footnote}]{The full name of \textit{Robot} dataset is \textit{``Wall-Following Robot Navigation Data''}}
\addtocounter{footnote}{1}
\footnotetext[\value{footnote}]{The full name of \textit{Statlog} dataset is \textit{``Statlog (Vehicle Silhouettes)''}}
\addtocounter{footnote}{1}
\footnotetext[\value{footnote}]{The full name of \textit{Vowel} dataset is \textit{``Connectionist Bench (Vowel Recognition - Deterding Data)''}}
\addtocounter{footnote}{1}
\footnotetext[\value{footnote}]{The full name of \textit{Steel} dataset is \textit{``Steel Plates Faults''}}
\section{Results}
\label{sec:res}
First, we investigate the performance of regularized learning of the weights with the hinge loss compared to the conventional least squares loss \cite{Reid_2009}  and the multi-response linear regression method which does not contain regularization \cite{ting99isg} with the diverse ensemble setup described in section \ref{sec:exp}. It should be noted that results shown here and in \cite{Reid_2009,ting99isg} are not directly comparable since construction of the ensembles is different. Error percentages of these three different learning algorithms for WS, CWS and LSG are given in Table \ref{table:mlr}. Results for the simple sum rule, which is equivalent to using equal weights in the WS, are also given in the column titled \textit{EW}. The first entries in the boxes are the means of error percentages over $5\times 2$ CV stacks and the second entries are the standard deviations.
\begin{table*}[htb]
\caption{Error percentages in the diverse ensemble setup (\textit{mean} $\pm$ \textit{standard deviation}).}
\centering
\scalebox{0.7}{
%\begin{tabular*}{1.0\columnwidth}{|c|c|c|c|c|c|c|c|c|c|}
\begin{tabular}{|c|c|c|c|c|c|c|c|c|c|c|}
%\scalebox{0.8}
\hline 
DB & \multicolumn{3}{|c|}{Hinge Loss with $l_2$ regularization} & \multicolumn{3}{|c|}{Least Squares Loss with $l_2$ regularization} & \multicolumn{3}{|c|}{MLR}& EW\\
         &   WS   &   CWS  &    LSG   &  WS    & CWS  &  LSG  &   WS   &   CWS  &    LSG   & \\
\hline \hline
 Segment & 5.02 $\pm$ 0.88  & 3.59 $\pm$ 0.96  & \textbf{3.44 $\pm$ 0.61}  & 6.44 $\pm$ 0.75  & 3.69 $\pm$ 0.89  & 3.59 $\pm$ 0.76  & 7.20 $\pm$ 1.02  & 6.66 $\pm$ 6.64  & 61.28 $\pm$ 9.35& 7.37 $\pm$ 1.03 \\
 \hline 
 Waveform & 13.20 $\pm$ 0.69  & 13.08 $\pm$ 0.76  & \textbf{13.05 $\pm$ 0.65}  & 13.15 $\pm$ 0.71  & 13.30 $\pm$ 0.76  & 13.15 $\pm$ 0.75  & 13.33 $\pm$ 0.68  & 14.10 $\pm$ 0.56  & 18.40 $\pm$ 7.06& 14.17 $\pm$ 0.60 \\
 \hline 
 Robot & 3.95 $\pm$ 0.42  & 2.53 $\pm$ 0.28  & 2.61 $\pm$ 0.28  & 5.21 $\pm$ 0.58  & 2.52 $\pm$ 0.30  & \textbf{2.50 $\pm$ 0.27}  & 5.05 $\pm$ 0.62  & 2.58 $\pm$ 0.30  & 3.19 $\pm$ 0.49& 18.58 $\pm$ 0.61 \\
 \hline 
 Statlog & 16.34 $\pm$ 1.15  & \textbf{16.12 $\pm$ 1.94}  & 16.36 $\pm$ 1.67  & 17.64 $\pm$ 1.65  & 16.90 $\pm$ 1.89  & 16.76 $\pm$ 1.64 & 17.73 $\pm$ 2.11  & 58.01 $\pm$ 15.38  & 75.72 $\pm$ 6.18& 23.03 $\pm$ 2.33 \\
 \hline 
 Vowel & 13.84 $\pm$ 2.73  & 6.79 $\pm$ 1.31  & \textbf{6.30 $\pm$ 1.99}  & 13.98 $\pm$ 2.64  & 6.55 $\pm$ 2.20  & 6.55 $\pm$ 1.85  & 17.15 $\pm$ 2.31  & 10.08 $\pm$ 1.75  & 9.76 $\pm$ 1.14& 14.53 $\pm$ 3.30 \\
 \hline 
 Wine & 2.13 $\pm$ 1.54  & \textbf{1.69 $\pm$ 1.52}  & 1.91 $\pm$ 1.76  & 2.70 $\pm$ 2.13  & 2.25 $\pm$ 1.83  & 6.85 $\pm$ 17.36 & 3.71 $\pm$ 2.31  & 8.20 $\pm$ 16.19  & 2.47 $\pm$ 1.66& 2.81 $\pm$ 1.52 \\
 \hline 
 Yeast & 40.36 $\pm$ 1.21  & 40.63 $\pm$ 1.21  & 40.70 $\pm$ 1.68  & 46.15 $\pm$ 18.76  & 40.42 $\pm$ 1.03  & 41.36 $\pm$ 1.32  & 41.05 $\pm$ 1.04  & 53.11 $\pm$ 6.88  & 74.45 $\pm$ 6.42  & \textbf{40.26 $\pm$ 1.10} \\
 \hline 
 Steel & 29.85 $\pm$ 1.86  & 27.37 $\pm$ 1.18  & 27.41 $\pm$ 1.22  & 31.06 $\pm$ 1.85  & \textbf{27.36 $\pm$ 1.17}  & 28.03 $\pm$ 2.77  &  30.35 $\pm$ 1.34   &  51.40 $\pm$ 14.66 & 77.12 $\pm$ 7.82 & 31.57 $\pm$ 2.07 \\
 \hline \hline
 MEAN & 15.59 $\pm$ 1.31  & \textbf{13.97 $\pm$ 1.15}  & \textbf{13.97 $\pm$ 1.23}  & 17.04 $\pm$ 3.63  & 14.12 $\pm$ 1.26  & 14.85 $\pm$ 3.34 & 16.94 $\pm$ 1.43  & 25.52 $\pm$ 7.80  & 40.30 $\pm$ 5.01 & 19.04 $\pm$ 1.57 \\
 \hline 
\end{tabular}}
\label{table:mlr} 
\end{table*}
For five datasets, the lowest error means are obtained with the hinge loss function and for two datasets lowest error means are obtained with the least-squares loss function. On \textit{yeast} dataset, simple averaging works better than the supervised learners. On all datasets, MLR method results in higher error percentages compared to other methods, and this shows the power of regularized learning, especially if the number of base classifiers is high. It should be noted that in \cite{ting99isg}, 3 base classifiers are used and here we use 130 base classifiers. According to the pairwise Wilcoxon signed-ranks test \cite{Demsar06}, hinge loss function outperforms least squares loss function at one-tailed significant level $\alpha = 0.05$ for WS and CWS combination types and at $\alpha=0.0525$ for LSG combination.

%If we compare different combination types, we see that there is no statistical significance between the errors of CWS and LSG. When we apply the significance test to each dataset independently, we see that there is a significant difference only on \textit{Robot} dataset. Considering 

\begin{figure}[!t]
\begin{center}
\includegraphics[width=1\linewidth]{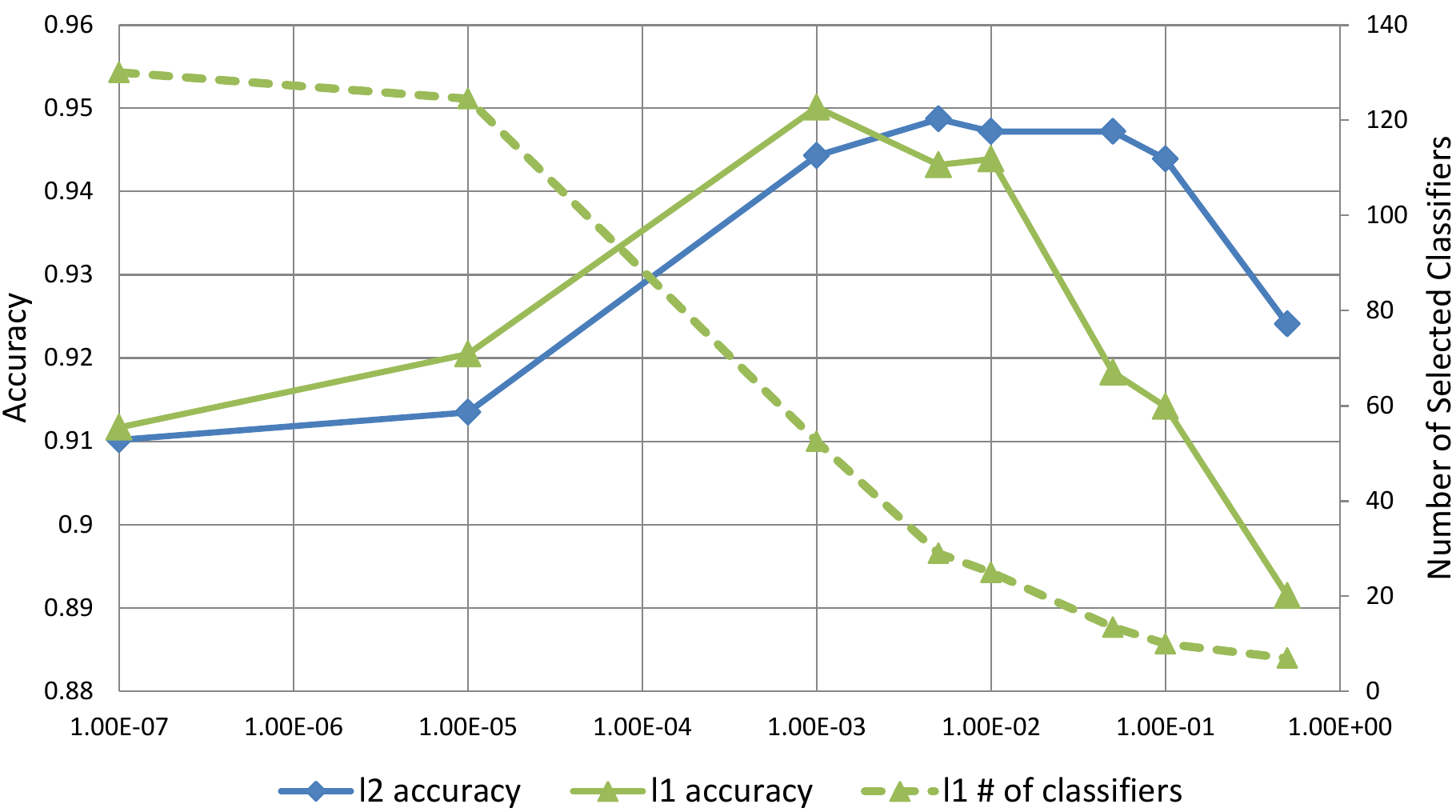}
\caption{\label{fig:type1} Accuracy and Number of selected classifiers vs. $\lambda$ for WS combination of Robot data with the diverse ensemble setup}
\end{center}
\end{figure}

\begin{figure}[!t]
\begin{center}
\includegraphics[width=1\linewidth]{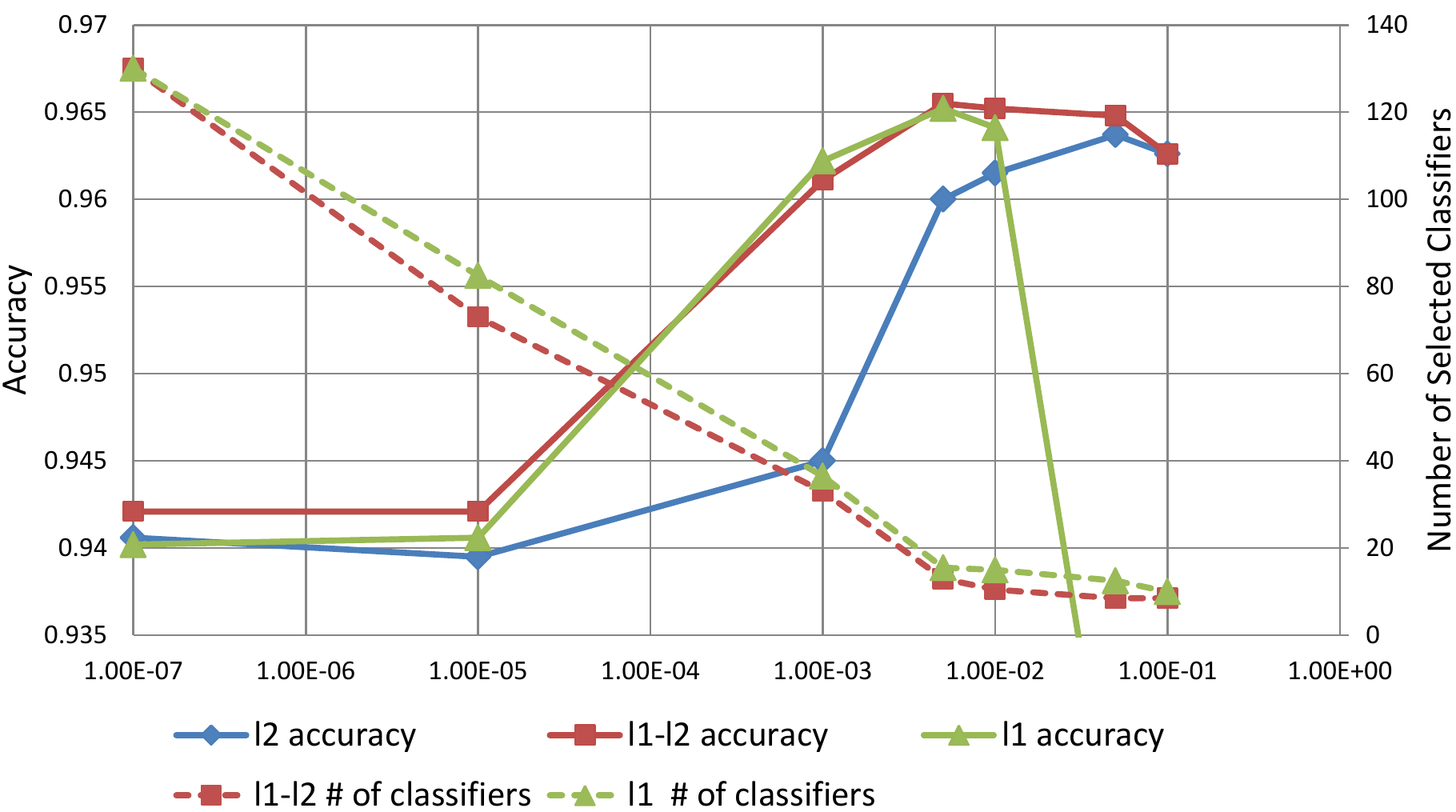}
\caption{\label{fig:type2} Accuracy and Number of selected classifiers vs. $\lambda$ for CWS combination of Robot data with the diverse ensemble setup}
\end{center}
\end{figure}

\begin{figure}[!t]
\begin{center}
\includegraphics[width=1\linewidth]{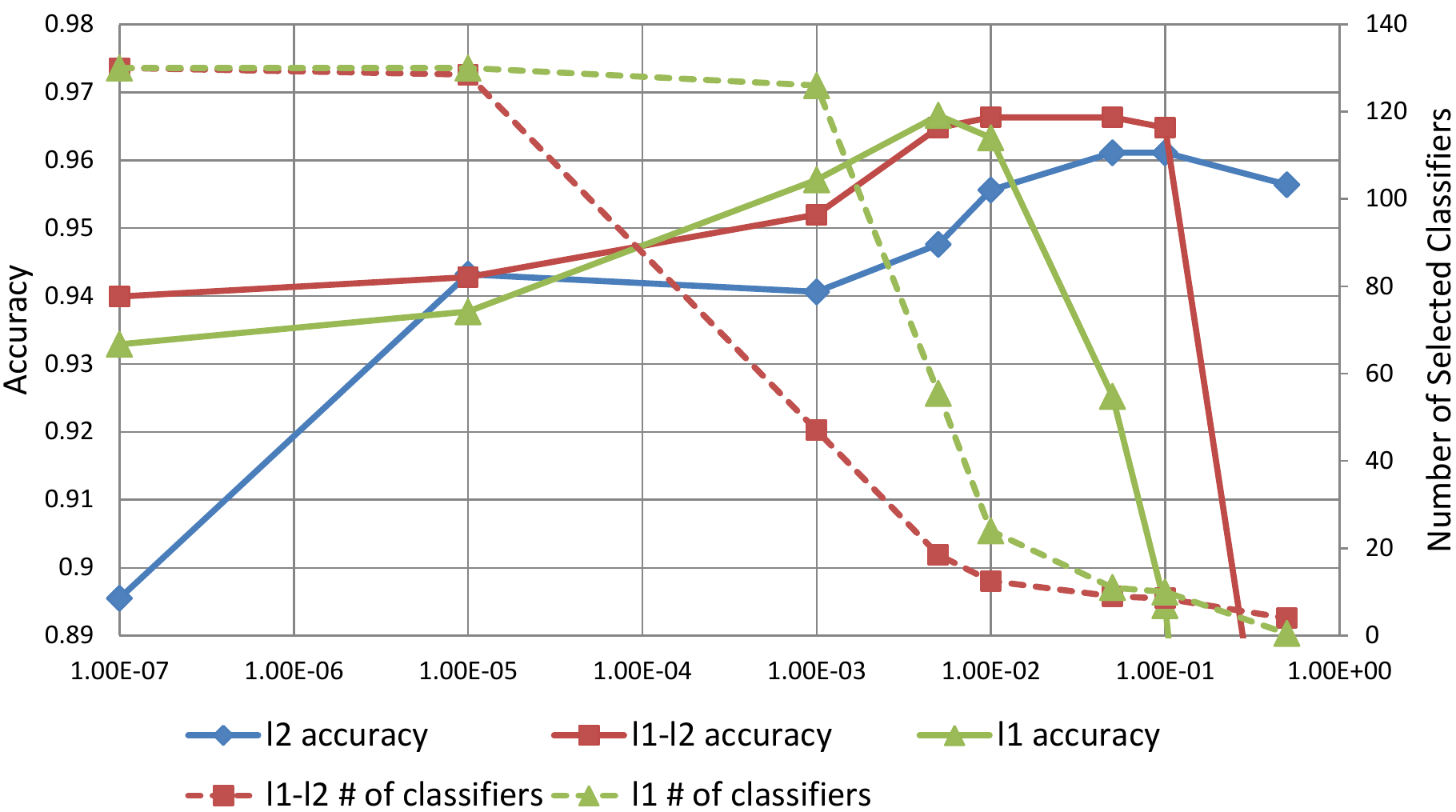}
\caption{\label{fig:type3} Accuracy and Number of selected classifiers vs. $\lambda$ for LSG combination of Robot data with the diverse ensemble setup}
\end{center}
\end{figure}

We also investigated the performance of sparse regularization with the hinge loss function. We used two different ensemble setups described in the beginning of this section. Regularization parameter $\lambda$ given in the objective functions (\ref{eq:rerm},\ref{eq:rermWS},\ref{eq:rermCWS}) is an important parameter and if we minimize the objective functions also over $\lambda$, the combiner will overfit the training data, which will result in poor generalization performance. Therefore, we used 2-fold cross-validation to learn the optimal parameter. We plot the relation of $\lambda$ with accuracies and the number of selected classifiers for different regularizations with WS, CWS and LSG for \textit{Robot} dataset in Figures \ref{fig:type1}, \ref{fig:type2} and \ref{fig:type3} respectively. 
In these figures, dashed lines correspond to the number of selected classifiers and solid lines correspond to accuracies. The $l_1-l_2$ label represents group sparsity. In all sparse regularizations, the best accuracies are obtained when most of the base classifiers are eliminated. For all regularizations, accuracies make a peak at $\lambda$ values between 0.001 and 0.1. For $l_1$ norm regularization, accuracies drop dramatically with a small increase in $\lambda$. However, with group sparsity regularization, accuracies remain high in a larger range for $\lambda$ than that with the $l_1$ norm regularization. Thus the performance of $l_1$ regularization is more sensitive to the selection of $\lambda$. So we can say that the $l_1-l_2$ norm regularization is more robust than $l_1$ norm regularization. As the number of selected classifiers decrease, accuracies increase in general, but this increase in the accuracy cannot be attributed to the classifier selection, because $\lambda$ also determines how much the combiner should fit the data as discussed in section \ref{sec:sparse}.

Next, we show the test results for all combination types with various regularization functions. Error percentages (mean $\pm$ standard deviation) are shown in Table \ref{table:scen1err} for the diverse ensemble setup and corresponding number of selected classifiers are shown in Table \ref{table:scen1noc}. 
\begin{table*}[htb]
\caption{Error percentages with the diverse ensemble setup (\textit{mean} $\pm$ \textit{standard deviation}). Bold values are the lowest error percentages of sparse regularizations ($l_1$ or $l_1-l_2$ regularizations)}
\centering
\scalebox{0.7}{
%\begin{tabular*}{1.0\columnwidth}{|c|c|c|c|c|c|c|c|c|c|}
\begin{tabular}{|c|c|c|c|c|c|c|c|c|c|}
%\scalebox{0.8}
\hline
DB & \multicolumn{2}{|c|}{WS} & \multicolumn{3}{|c|}{CWS} & \multicolumn{3}{|c|}{LSG} & EW\\
   & $l_2$ & $l_1$ & $l_2$ & $l_1$ & $l_1-l_2$ & $l_2$ & $l_1$ & $l_1-l_2$&\\
	\hline \hline
 Segment & 5.02 $\pm$ 0.88  & 4.90 $\pm$ 0.99  & 3.59 $\pm$ 0.96  & 3.62 $\pm$ 0.62  & 3.74 $\pm$ 0.40  & 3.44 $\pm$ 0.61  & 3.79 $\pm$ 1.05  & \textbf{3.29 $\pm$ 0.55}  & 7.37 $\pm$ 1.03 \\
 \hline 
 Waveform & 13.20 $\pm$ 0.69  & 13.38 $\pm$ 0.70  & 13.08 $\pm$ 0.76  & 13.46 $\pm$ 0.74  & 13.42 $\pm$ 0.76  & 13.05 $\pm$ 0.65  & 13.33 $\pm$ 0.71  & \textbf{13.24 $\pm$ 0.64 } & 14.17 $\pm$ 0.60 \\
 \hline 
 Robot & 3.95 $\pm$ 0.42  & 4.00 $\pm$ 0.38  & 2.53 $\pm$ 0.28  & 2.57 $\pm$ 0.35  & \textbf{2.49 $\pm$ 0.33}  & 2.61 $\pm$ 0.28  & 2.54 $\pm$ 0.35  & 2.52 $\pm$ 0.32  & 18.58 $\pm$ 0.61 \\
 \hline 
 Statlog & 16.34 $\pm$ 1.15  & \textbf{17.19 $\pm$ 1.63}  & 16.12 $\pm$ 1.94  & 17.45 $\pm$ 1.74  & 17.33 $\pm$ 1.42  & 16.36 $\pm$ 1.67  & 17.40 $\pm$ 1.34  & 17.45 $\pm$ 1.51  & 23.03 $\pm$ 2.33 \\
 \hline 
 Vowel & 13.84 $\pm$ 2.73  & 14.40 $\pm$ 2.27  & 6.79 $\pm$ 1.31  & 7.62 $\pm$ 2.02  & 7.17 $\pm$ 1.50  & 6.30 $\pm$ 1.99  & \textbf{6.18 $\pm$ 1.19}  & 6.79 $\pm$ 1.17  & 14.53 $\pm$ 3.30 \\
 \hline 
Wine & 2.13 $\pm$ 1.54  & 2.13 $\pm$ 1.63  & 1.69 $\pm$ 1.52  & 2.25 $\pm$ 1.18  & \textbf{1.91 $\pm$ 1.30}  & 1.91 $\pm$ 1.76  & 2.25 $\pm$ 1.59  & 2.36 $\pm$ 1.54  & 2.81 $\pm$ 1.52 \\
 \hline 
 Yeast & 40.36 $\pm$ 1.21  & \textbf{40.38 $\pm$ 1.06}& 40.63 $\pm$ 1.21  & 42.53 $\pm$ 4.42  & 41.19 $\pm$ 1.57  & 40.70 $\pm$ 1.68  & 48.09 $\pm$ 18.30  & 41.67 $\pm$ 1.31  & 40.26 $\pm$ 1.10 \\
 \hline 
 Steel & 29.85 $\pm$ 1.86  & 30.00 $\pm$ 2.61  & 27.37 $\pm$ 1.18  & 28.31 $\pm$ 1.39  & \textbf{27.41 $\pm$ 1.21}  & 27.41 $\pm$ 1.22  & 28.09 $\pm$ 1.03  & 27.50 $\pm$ 1.24  & 31.57 $\pm$ 2.07 \\
 \hline \hline
 MEAN & 15.59 $\pm$ 1.31  & 15.80 $\pm$ 1.41  & 13.97 $\pm$ 1.15  & 14.73 $\pm$ 1.56  & \textbf{14.33 $\pm$ 1.06}  & 13.97 $\pm$ 1.23  & 15.21 $\pm$ 3.20  & 14.35 $\pm$ 1.04  & 19.04 $\pm$ 1.57 \\
 \hline 
\end{tabular}}
\label{table:scen1err} 
\end{table*}
\begin{table*}[htb]
\caption{Number of selected classifiers with the diverse ensemble setup out of 130 (\textit{mean} $\pm$ \textit{standard deviation}).}
\centering
\scalebox{0.7}{
\begin{tabular}{|c|c|c|c|c|c|}
\hline 
DB & WS  & \multicolumn{2}{|c|}{CWS} & \multicolumn{2}{|c|}{LSG}\\
   &  $l_1$ &  $l_1$ & $l_1-l_2$ &  $l_1$ & $l_1-l_2$\\
	\hline \hline
Segment & \textbf{21.50 $\pm$ 4.62}  & 63.50 $\pm$ 25.72  & 30.80 $\pm$ 34.92  & 97.40 $\pm$ 24.40  & 80.40 $\pm$ 14.93 \\
 \hline 
 Waveform & 36.60 $\pm$ 49.44  & 23.30 $\pm$ 37.59  & 47.00 $\pm$ 57.31  & \textbf{11.20 $\pm$ 2.30 } & 12.10 $\pm$ 5.38 \\
 \hline 
Robot  & 41.80 $\pm$ 9.02  & 18.60 $\pm$ 5.97  & 14.00 $\pm$ 4.55  & 18.50 $\pm$ 4.53  & \textbf{13.30 $\pm$ 2.63} \\
 \hline 
 Statlog & 36.10 $\pm$ 34.75  & 14.30 $\pm$ 10.85  & 49.20 $\pm$ 56.13  & 30.60 $\pm$ 36.31  &\textbf{ 11.20 $\pm$ 12.42} \\ 
 \hline 
Vowel & 108.90 $\pm$ 44.48  & 37.80 $\pm$ 32.62  & 57.30 $\pm$ 62.64  & 128.00 $\pm$ 6.32  & \textbf{13.80 $\pm$ 3.99} \\ 
 \hline 
 Wine & 130.00 $\pm$ 0.00  & 121.30 $\pm$ 18.60  & 117.10 $\pm$ 40.44  & 93.50 $\pm$ 58.86  & \textbf{91.60 $\pm$ 61.83} \\ 
 \hline 
 Yeast & 119.10 $\pm$ 34.47  & 121.00 $\pm$ 28.46  & 40.40 $\pm$ 47.33  & 130.00 $\pm$ 0.00  & \textbf{9.80 $\pm$ 3.46} \\
 \hline 	
 Steel & 41.90 $\pm$ 32.05  & 42.10 $\pm$ 6.85  & 35.30 $\pm$ 8.10  & 51.00 $\pm$ 16.62  & \textbf{35.20 $\pm$ 11.93} \\
 \hline \hline
 MEAN & 66.99 $\pm$ 26.10  & 55.24 $\pm$ 20.83  & 48.89 $\pm$ 38.93  & 70.03 $\pm$ 18.67  & \textbf{33.43 $\pm$ 14.57} \\
\hline
\end{tabular}}
\label{table:scen1noc} 
\end{table*}

In general, we are able to use much less base classifiers with sparse regularizations with the cost of a small decrease in the accuracies. For CWS, group sparsity regularization outperforms $l_1$ norm regularization at one-tailed significance level $\alpha = 0.005$. For LSG, average error percentage of group sparsity is a little less than that of the $l_1$ norm regularization which is not statistically significant. But the number of selected base classifiers is much less. So if classifier selection is desired, we suggest to use either CWS or LSG combination with $l_1-l_2$ regularization. If training time is also crucial, CWS with $l_1-l_2$ regularization seems to be the best option.
%If we compare different combination types with the $l_2$ norm, we see that LSG gives the lowest error and WS gives the highest error. So, in general, there is a trade-off between accuracy and training time for the diverse ensemble, since training time of WS is less than that of CWS and CWS's training time is also less than that of LSG. If we compare different sparse regularization results, we see that LSG with $l_1-l_2$ norm regularization gives the almost lowest mean error with smallest standard deviations and with smallest number of selected classifiers. So if classifier selection is desired for a classifier combination system, LSG with $l_1-l_2$ regularization seems to be the best option. If training time is also crucial, CWS with $l_1-l_2$ norm regularization can be preferred.

Error percentages and number of selected classifiers for the non-diverse ensembles are given in Tables \ref{table:scen2err} and \ref{table:scen2noc} respectively.
\begin{table*}[htb]
\caption{Error percentages with the non-diverse ensemble setup(\textit{mean} $\pm$ \textit{standard deviation}). Bold values are the lowest error percentages of sparse regularizations ($l_1$ or $l_1-l_2$ regularizations).}
\centering
\scalebox{0.7}{
%\begin{tabular*}{1.0\columnwidth}{|c|c|c|c|c|c|c|c|c|c|}
\begin{tabular}{|c|c|c|c|c|c|c|c|c|c|}
%\scalebox{0.8}
\hline 
DB & \multicolumn{2}{|c|}{WS} & \multicolumn{3}{|c|}{CWS} & \multicolumn{3}{|c|}{LSG} & EW\\
   & $l_2$ & $l_1$ & $l_2$ & $l_1$ & $l_1-l_2$ & $l_2$ & $l_1$ & $l_1-l_2$&\\
	\hline \hline
Segment & 4.35 $\pm$ 0.66  & 4.49 $\pm$ 0.71  & 4.30 $\pm$ 0.71  & \textbf{4.21 $\pm$ 0.80}  & 4.33 $\pm$ 0.74  & 5.11 $\pm$ 0.90  & 9.78 $\pm$ 17.11  & 4.35 $\pm$ 0.75  & 11.37 $\pm$ 0.74 \\
 \hline 
 Waveform & 13.23 $\pm$ 0.73  & \textbf{13.12 $\pm$ 0.81}  & 13.30 $\pm$ 0.80  & 13.33 $\pm$ 0.75  & 13.24 $\pm$ 0.78  & 13.22 $\pm$ 0.76  & 13.20 $\pm$ 0.70  & 13.25 $\pm$ 0.68  & 13.34 $\pm$ 0.80 \\
 \hline 
 Robot & 7.99 $\pm$ 0.66  & 7.98 $\pm$ 0.70  & 7.94 $\pm$ 0.63  & 8.13 $\pm$ 0.40  & \textbf{7.94 $\pm$ 0.49}  & 8.01 $\pm$ 0.70  & 8.13 $\pm$ 0.55  & 7.99 $\pm$ 0.56  & 10.70 $\pm$ 0.47 \\
 \hline 
 Statlog & 18.65 $\pm$ 2.18  & 18.87 $\pm$ 2.05  & 18.49 $\pm$ 1.76  & \textbf{18.77 $\pm$ 1.74}  & 19.24 $\pm$ 1.81  & 19.62 $\pm$ 1.15  & 19.17 $\pm$ 2.17  & 19.10 $\pm$ 1.56  & 28.32 $\pm$ 1.82 \\
 \hline 
 Vowel & 6.91 $\pm$ 2.34  & 9.88 $\pm$ 3.46  & 8.04 $\pm$ 2.12  & 6.34 $\pm$ 2.29  & \textbf{6.08 $\pm$ 2.37}  & 8.57 $\pm$ 2.10  & 7.72 $\pm$ 2.24  & 6.10 $\pm$ 2.30  & 20.16 $\pm$ 2.74 \\
 \hline 
 Wine & 9.21 $\pm$ 1.82  & 8.88 $\pm$ 2.45  & 8.76 $\pm$ 1.97  & 8.65 $\pm$ 3.05  & \textbf{8.20 $\pm$ 3.63 } & 14.94 $\pm$ 9.22  & 8.65 $\pm$ 2.31  & 8.99 $\pm$ 2.95  & 27.75 $\pm$ 7.57 \\
 \hline  \hline
 MEAN & 10.06 $\pm$ 1.40  & 10.54 $\pm$ 1.70  & 10.14 $\pm$ 1.33  & 9.91 $\pm$ 1.51  & \textbf{9.84 $\pm$ 1.64}  & 11.58 $\pm$ 2.47  & 11.11 $\pm$ 4.18  & 9.96 $\pm$ 1.47  & 18.61 $\pm$ 2.36 \\
 \hline  
\end{tabular}}
\label{table:scen2err} 
\end{table*}
\begin{table*}[htb]
\caption{Number of selected classifiers out of 154 with the non-diverse ensemble setup.}
\centering
\scalebox{0.7}{
\begin{tabular}{|c|c|c|c|c|c|}
\hline 
DB & WS  & \multicolumn{2}{|c|}{CWS} & \multicolumn{2}{|c|}{LSG}\\
   &  $l_1$ &  $l_1$ & $l_1-l_2$ &  $l_1$ & $l_1-l_2$\\
	\hline \hline
 Segment & 29.40 $\pm$ 8.13  & 13.60 $\pm$ 6.93  & 8.40 $\pm$ 6.93  & 51.80 $\pm$ 70.80  & \textbf{2.60 $\pm$ 2.67} \\ 
 \hline 
 Waveform & 32.40 $\pm$ 64.10  & 51.60 $\pm$ 70.98  & 95.60 $\pm$ 75.54  & \textbf{5.10 $\pm$ 3.18}  & 36.70 $\pm$ 62.37 \\ 
 \hline 
 Robot & 43.80 $\pm$ 16.19  & 30.60 $\pm$ 10.20  & 28.30 $\pm$ 16.57  & 20.50 $\pm$ 15.71  & \textbf{13.70 $\pm$ 3.40} \\ 
 \hline 
 Statlog & 18.90 $\pm$ 9.46  & 14.50 $\pm$ 11.98  & 8.90 $\pm$ 9.64  & 25.30 $\pm$ 46.07  & \textbf{7.80 $\pm$ 5.03} \\ 
 \hline 
 Vowel & 69.20 $\pm$ 59.27  & 8.70 $\pm$ 10.12  & 3.00 $\pm$ 2.83  & 125.70 $\pm$ 45.82  & \textbf{1.40 $\pm$ 0.70} \\ 
 \hline 
 Wine & 65.00 $\pm$ 73.24  & 39.30 $\pm$ 60.96  & \textbf{21.90 $\pm$ 46.09}  & 34.80 $\pm$ 62.84  & 78.90 $\pm$ 79.19 \\ 
 \hline \hline 
 MEAN & 43.12 $\pm$ 38.40  & 26.38 $\pm$ 28.53  & 27.68 $\pm$ 26.27  & 43.87 $\pm$ 40.74  & \textbf{23.52 $\pm$ 25.56} \\ 
 \hline
\end{tabular}}
\label{table:scen2noc} 
\end{table*}
With the non-diverse ensembles we are even able to increase the accuracy with much less number of base classifiers with sparse regularization in CWS and LSG. On the average, $l_1-l_2$ regularization results in lower error percentages for both CWS and LSG, but the results are not statistically significant. But, the number of selected classifiers is much less with $l_1-l_2$ regularization than that of $l_1$ regularization. Except \textit{statlog} dataset, lowest error percentages are obtained with the sparse combinations with much less base classifiers than that of $l_2$ regularization which uses 154 base classifiers. If we compare different combination types with the $l_2$ norm, on the average we see that, unlike in the diverse ensemble setup, WS and CWS outperforms LSG in four databases. We can conclude that if the posterior scores obtained from base classifiers are correlated, non-complex combiners are more powerful since complex combiners may result in overfitting.

\section{Conclusion}
In this paper, we suggested using hinge loss function with regularization to learn the parameters (or weights) of linear combiners in stacked generalization. We are able to obtain better accuracies with the hinge loss function than conventional least-squares estimation of the weights. Results also indicate the importance of the regularized learning of the weights. We also proposed $l_1-l_2$ norm regularization (or group sparsity) to obtain a reduced number of base classifiers so that the test time is shortened. Results indicate that we can use smaller number of base classifiers with a small sacrifice in the accuracy with the diverse ensemble. We show that $l_1-l_2$ regularization outperforms $l_1$ regularization in terms of both accuracy and the number of selected classifiers. With the non-diverse ensemble setup, we even obtain better accuracies using sparse regularizations. If training time is crucial, we suggest using CWS type combination. And if test time is also important, we suggest using group sparsity regularization.

\section{Acknowledgments}
This research is supported by The Scientific and Technological Research Council of Turkey (TUBITAK) under the scientific and technological research support program (code 1001), project number 107E015 entitled ``Novel Approaches in Audio Visual Speech Recognition''.

\bibliographystyle{IEEEtran}
\bibliography{IEEEabrv,classiferCombination}
\end{document}